# Service-Oriented Architecture for Space Exploration Robotic Rover Systems

Youssef Bassil

*LACSC – Lebanese Association for Computational Sciences*
*Registered under No. 957, 2011, Beirut, Lebanon*
`youssef.bassil@lacsc.org`

*Abstract*—Currently, industrial sectors are transforming their business processes into e-services and component-based architectures to build flexible, robust, and scalable systems, and reduce integration-related maintenance and development costs. Robotics is yet another promising and fast-growing industry that deals with the creation of machines that operate in an autonomous fashion and serve for various applications including space exploration, weaponry, laboratory research, and manufacturing. It is in space exploration that the most common type of robots is the planetary rover which moves across the surface of a planet and conducts a thorough geological study of the celestial surface. This type of rover system is still ad-hoc in that it incorporates its software into its core hardware making the whole system cohesive, tightly-coupled, more susceptible to shortcomings, less flexible, hard to be scaled and maintained, and impossible to be adapted to other purposes. This paper proposes a service-oriented architecture for space exploration robotic rover systems made out of loosely-coupled and distributed web services. The proposed architecture consists of three elementary tiers: the client tier that corresponds to the actual rover; the server tier that corresponds to the web services; and the middleware tier that corresponds to an Enterprise Service Bus which promotes interoperability between the interconnected entities. The niche of this architecture is that rover's software components are decoupled and isolated from the rover's body and possibly deployed at a distant location. A service-oriented architecture promotes integrate-ability, scalability, reusability, maintainability, and interoperability for client-to-server communication. Future research can improve upon the proposed architecture so much so that it supports encryption standards so as to deliver data security as well as message concealment for the various communicating entities of the system.

## 1. Introduction

Robotics technology is emerging at a rapid pace, offering new possibilities for automating tasks in many challenging applications, especially in space explorations, military operations, underwater missions, domestic services, and medical procedures. Particularly, in space exploration, robotic devices are formally known as planetary rovers or simply rovers and they are aimed at conducting physical analysis of planetary terrains and astronomical bodies, and collecting data about air pressure, climate, temperature, wind, and other atmospheric phenomena surrounding the landing sites [1]. Basically, rovers can be autonomous capable of operating with little or no assistance from ground control or they can be remotely controlled from earth ground stations called RCC short for Remote Collaboration Center. In practice, robotic rovers have very definite scientific objectives [2]. They include but not limited to the examining of territories at the microscopic level, carrying out physical experimentations, investigating the biological aspects of planetary surfaces, analyzing the composition of minerals, rocks, and soils, searching for geological clues such as finding liquid water in minerals, and measuring the ambient temperature, air pressure, and amount of dust in the landing site [3]. As a result, rovers are highly computing intensive systems that use complex embedded software and algorithms to handle computational and processing tasks. These days, it is no longer practical to develop ad-hoc systems that require a single person to wisely craft the entire software for the rover. In addition, it is no more feasible to encapsulate all software components within the actual rover hardware. Instead, a component-based model or service-oriented architecture is often followed in which the rover's software is developed as a set of services by multiple persons working on a large code base in a distributed team [4].

Inherently, a service is a software component that contains a collection of related software functionalities reusable for different purposes [5]. It delivers such operations as data storage, data processing, mathematical and scientific computations, and networking. It is governed by a producer-consumer model in which a service is delivered by a service provider known as the producer which owns the equipment for hosting, running, and maintaining the service, and the client known as the consumer which connects and uses service functionalities via remote method invocation mechanism. Predominantly, services are implemented as Web Services (WS) which are defined by the W3C as "a software system designed to support interoperable machine-to-machine interaction over a network" [6].

This paper proposes a service-oriented architecture for autonomous space exploration robotic rover systems based on heterogeneous multi-platform service components. The proposed architecture is composed of three basic tiers: The first tier is the client represented by the actual robotic rover vehicle. The second tier is the server which hosts and runs the different service components that provide the advanced functionalities necessary for the rover operations. Services are decoupled and isolated from the actual rover and possibly deployed at a distant location. The third tier is the middleware represented by an Enterprise Service Bus (ESB) which offers a standard interface and a data-path for both the client and server tiers to interact, send requests, and receive responses from each other.

Being decentralized and decoupled from the rover's hardware core, the proposed service-oriented architecture has five benefits [7][8][9]: Integrate-ability which allows the seamless integration of new software components in a less significant effort, time, and budget; reusability which is given by the nature of SOA "build once, use many times" that allows multiple rovers, possibly located in different sites, to use and share the same set of services simultaneously and with high availability; scalability which is given by the ability to add, update, and delete rover's functionalities remotely with no or minimal service interruption and while the system is online; maintainability which is given by that a failure in a service would only require replacing the faulty service and not the entire rover system; and interoperability which is given by the Enterprise Service Bus middleware which provides a standardized and a unified platform for the various interconnected entities, possibly incompatible, to send and receive data among each other.

## 2. Space Exploration Rovers

In essence, a rover is a space exploration robotic vehicle used particularly in exploring the land of a planet. It has the capability to travel across the surface of a landscape and other cosmic bodies. A rover has many features: It can generate power from solar panels; capture high-resolution images; move in 360 degrees with the help of a navigation camera (Navcam); walk across obstacles such as bumps and rocks; conduct deep analysis and record measurements using multiple types of spectrometers; find properties of materials to identify their types and their composition; search for geological clues such as water to detect any presence of life on the landing environment; and inspect the mineralogy and texture of the local terrain using panoramic cameras (Pancam) [10][11].

There exist two types of rover vehicles: The first type is the human-controlled rovers which are remotely manipulated from earth and usually guided to perform a particular operation. Communication between the rover and the earth control occurs through the Deep Space Network (DSN), which is an international network of large antennas with communication facilities that supports interplanetary spacecraft missions. Currently, DSN comprises three deep-space communications facilities located in Mojave Desert in California, west of Madrid in Spain, and south of Canberra in Australia. The second type is the autonomous rovers which can complete desired tasks without constant human direction. Space exploration rovers are distinguished by a high degree of autonomy as they can cope with their changing environment, automatically gain information about the landing sites, survive a disaster or a failure, operate for prolonged periods of time, and execute predefined operations without human assistance. Financially, robotic rovers can cost to build, test, and deploy hundreds of millions of dollars sometimes billions of dollars [12]. Historically, Lunokhod and Marsokhod were two space rovers designed and launched by the soviet in the 70s [13]; while, Spirit and Opportunity were two US rovers produced by NASA, the space agency of the United States, between year 2004 and 2010 as part of NASA's ongoing Mars Exploration Rover Mission (MER).

From a design point of view, the brain of the rover is contained in what so called WEB short for Warm Electronic Box [14]. Traditional rover architectures use the WEB to house two basic entities: a high-end digital computer able to perform computations at very high speed and a software which controls the rover's hardware and provides all its required functionalities. On the other hand, a service-oriented architecture would decouple the software from the WEB of the rover and place it into a server possibly located on earth or in space station operated in low earth orbit that can communicate with the rover remotely to deliver its necessary functionalities. Additionally, using a service-oriented architecture, the rover is no more a monocoque system made out of one single unit but of loosely-coupled distributed components that are separated from its physical hardware and hosted in a remote location.

## 3. Service-Oriented Architecture

Service-Oriented Architecture or SOA for short is a model for system development based on loosely-integrated suite of services that can be used within multiple business domains [15]. SOA is also an approach and practice for building IT software systems using interoperable services. These services are loosely-coupled software components that encapsulate functionalities and are available to be remotely accessed by client applications over a network or Internet [16]. The backbone of SOA consists of web services and an Enterprise Service Bus (ESB).

### 3.1 Web Services

As defined by W3C, a web service is a software component designed to support interoperable machine-to-machine interaction over a network [6]. It uses the SOAP, an XML-based protocol to communicate over HTTP. Characteristically, web services have three key elements: Web Service Description Language (WSDL) which is an XML-based description of the operations and functionalities offered by the web service. It dictates the protocol bindings and the message formats required to connect to and interact with a given web service; Universal Description, Discovery and Integration (UDDI) which is a registry for storing web services' WSDLs and a mechanism to register and locate web services on the Internet; and the SOAP communication protocol which defines the structure and format of the messages being exchanged between the service requester represented by the client and the service provider represented by the actual web service. In fact, the service requester is a client application requesting a particular functionality from the service provider, and the service provider is usually a server that hosts and runs the actual web service. Figure 1 illustrates the operation mode of a generic SOAP-based web service.

Other types or styles exist for web services. They include REST, RPC, RMI, .NET Remoting, CORBA, and Network Socket [17].

REST (Representational State Transfer) web services do not use the SOAP protocol to communicate; rather, they use the plain HTTP protocol and Query String information to exchange messages. Their advantages over SOAP-based web services are that they are easier to build, manage, and reuse.

RPC (Remote Procedure Call) is an inter-process communication that allows a computer program to invoke or call remotely a function or procedure to execute on another computer over a shared network. RMI (Remote Method Invocation) is the Java implementation for RPC, while .NET Remoting is the .NET implementation for RPC.

Network Socket is an inter-process communication between two or more computer programs over a network. A server socket uses a socket address which is a combination of an IP address and a port number to listen for incoming connections. Clients connect to the server socket and then start exchanging data packets. Network sockets can be implemented using either TCP or UDP protocols.

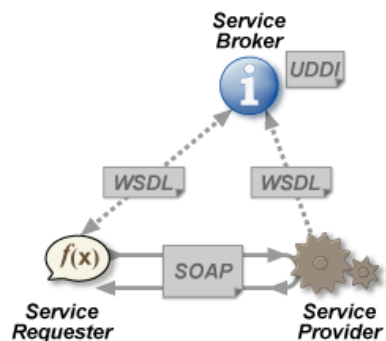

**Figure 1.** Typical SOAP-Based Web Service

### 3.2 Enterprise Service Bus - ESB

In order to promote interoperability among its components, SOA often employs an Enterprise Service Bus or ESB. Fundamentally, an ESB is a piece of software that lies between the different components of an SOA, mainly between the service requester and the service provider to enable a transparent and seamless communication among them [18]. It, in fact, acts as a middleware and a message broker between the different communicating parties in SOA architecture. The primary task of ESB is to support message routing and ensure a better orchestration and interoperability between the various interconnected web services possibly built using different technologies, platforms, standards, and programming languages.

## 4. Related Work

There are a large variety of different architectures, models, and frameworks already developed for robotic systems. Several of them are examined in this section:

One of the earliest models is the 3T robotic design [19] which is composed of three independent communicating layers: The first layer is the reactive skills layer which consists of a set of reactive behaviors handled by a skill manager which is responsible for starting and stopping particular skills based on several input sensors. The second layer is the sequencing layer which is responsible for generating a set of skill requirements for the reactive skills layer. The third layer is the deliberative layer which is responsible for generating plans based on mission requirements. The 3T software architecture was designed to effectively control autonomous systems in a flexible and robust manner.

Player [20] is a distributed robotic control protocol description that uses socket communications to convey its purpose. It uses POSIX network socket building blocks to deliver a transparent communication between the robot system and its various sensors. Player is universal as it relies on standard sockets; and thus can be implemented by any programming environment.

JAUS [21] is yet another architecture for robotic systems designed by the U.S. Department of Defense to support all its autonomous systems. In effect, JAUS is a component-based messaging system made out of modular programming units whose task is to control the flow of data in and out of the system. JAUS uses a strict messaging format which limits the integration of new components that are not JAUS-compatible.

Joint Technical Architecture [22] was developed by the U.S. Department of Defense to provide interoperable platform for robotic systems. JTA primarily focused on military machinery such as aerial systems, firing ground systems, and war fighting machinery.

Mobile and Autonomous Robotic Integration Environment or MARIE for short [23] was designed with the purpose of quickening and facilitating the integration of robotic systems. MARIE is based on software integration of loosely coupled distributed software components and provides a central mediator to control the integration of new components regardless of their underlying architectures and communication protocols.

Controlling Robots with CORBA also known as CoRoBa [24] is a development platform designed to permit the integration of distributed robotic control, sensor, and computational components. It is fundamentally based on CORBA (Common Object Request Broker Architecture) which provides a standard middleware for connecting object-oriented components together with little regards to their inner technologies.

Mobility Integration Architecture [25] is yet another CORBA-based software architecture for building and integrating distributed object systems built using different languages.

ROS which stands for Robot Operating System [26] is an open source architecture that supports modular and distributed software components for robotic software. The ROS architecture features interaction between entities, message passing, and services concept. The ROS design does not however feature a central middleware to coordinate among the different nodes of the system.

[27] proposed a service-oriented architecture for distributed multi-robot systems based on web services for realizing remote controlling and on manufacturing message specification (MMS) to exchange data among the different modules of the system. Its purpose is to monitor and control software design using Unified Modeling Language (UML) and MMS concepts and to enable such processes as e-manufacturing, e-diagnostics, and e-maintenance.

## 5. Proposed Architecture

This paper proposes a Service-Oriented Architecture (SOA) for building space exploration robotic rover machines using web service software components. It is a distributed model made out of loosely-coupled interoperable web services and a central Enterprise Service Bus (ESB) not located inside the actual rover but in an isolated location, possibly earth control center or space station in low earth orbit. The communication between the rover and the web services is bi-directional and is done in a remote fashion using the HTTP protocol with the help of the ESB acting as a middleware. The employed communication style is method invocation in which the rover remotely calls or invokes the different procedures of the web services to execute on the hosting system and return results to the rover. These procedures also known as methods or functions contain the logic and the programming instructions that deliver the rover's basic functionalities.

Essentially, the proposed architecture is composed of three basic tiers: The first tier is the client represented by the rover vehicle which invokes methods of web services to perform operations such as examining minerals, analyzing geological environment, studying and assessing the composition of rocks, and capturing and processing images for a variety of applications. The second tier is the server represented by the web services which are decoupled from the rover hardware and hosted and executed on server machines located on earth or in a nearby space station. The web services provide the actual code base and logic for the rover. They contain the algorithm, implementation, and programming instructions necessary to provide the rover its basic operations and functionalities. The third and final tier is the middleware represented by the Enterprise Service Bus which offers a standard interface and a unified data-path for both the client and the server tiers to interoperate efficiently and exchange data regardless of their incompatible platforms and implementation technologies, for instance, technologies such as SOAP, REST, RPC or others. Figure 2 illustrates the proposed SOA architecture and its different tiers.

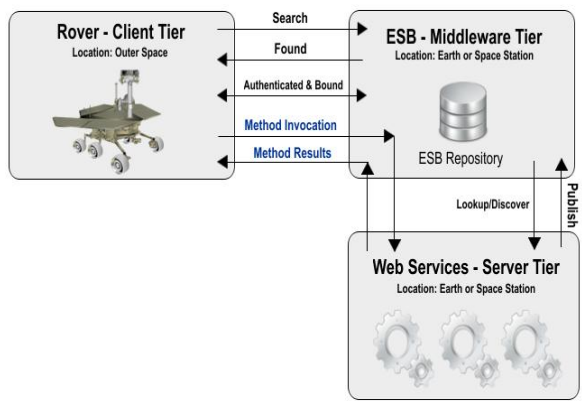

**Figure 2.** Tiers of the proposed SOA architecture

### 5.1 Advantages & Motivations

A service-oriented architecture for robotic rover systems would decouple and detach the rover's software from the core of the rover, making it independent and not physically bound to the actual rover's hardware. As a result, the rover is no more a single block housing both hardware and software; rather, only hardware is embedded within the rover, while the software consists of loosely-coupled distributed web services encapsulating the rover's basic functionalities and executed remotely outside the rover. The rover only sends requests to and gets results from web services. Such design has many advantages which can be listed below:

Integrate-ability: Integration of new software components can take less significant effort, time, and budget. For instance, new services providing new functionalities can be easily deployed on the server tier without the need to access the out of reach client rover. Likewise, changes to the existing web services can be easily made by only changing the service description on the server side.

Scalability: SOA is an open architecture in that it supports plug-and-play operations. For instance, new services can be deployed at runtime with no or minimal amount of system interruption. Similarly, they can be pulled out of the system at any time without experiencing degradation in performance or shortcomings in system operation. On the other hand, existing services can be reconfigured and updated at minimal cost. As SOA is governed by the publish-discover process [28], delivering new services and consuming them is usually done in an automated manner.

Maintainability: Since services are no more part of the rover and thus located at a great distance away from the landing sites, it is less tedious and less costly to isolate system defects and troubleshoot, diagnose, and repair broken services. Consequently, this promotes an agile and robust system that can cope with an unpredictable and always changing environment without affecting the system in operation.

Reusability: Services can be reused to add or extend new functionalities or build new rover systems from already existing components. This practice can reduce design, development, implementation, testing, and deployment time.

Decentralization: Being modular, SOA components can be dispersed over multiple hosting environments providing computing power over distributed and inexpensive machines of massive computing arrays.

Interoperability: As SOA features an ESB which emulates a middleware that sits between the rover and the web services, it provides a standardized and cross-network platform over which the rover can interoperate transparently with the different heterogeneous web services built using different standards, programming languages, technologies, and platforms.

## 5.2 Design Specifications

As discussed earlier, the proposed architecture comprises three tiers: The client, the middleware, and the server tier.

### 5.2.1 The Client Tier – The Rover Vehicle

Actually, the client is the robotic rover vehicle. It contains an onboard computer able to discover the different remote web services through the ESB interface which describes the different functions encapsulated within the connected web services. In order to communicate, the rover has to bind to the ESB interface. This binding authenticates the rover (requester) and allows it to send requests to the ESB (provider) using remote procedure invocation approach. The ESB then forwards the rover's request to the intended web service. The results that are returned by the web service are first received by the ESB then forwarded to the rover. All execution is done on the provider's side and only results are returned to the requester. Communication between requester and provider is done solely using the HTTP protocol through the Deep Space Network (DSN) that relays transmission between the earth where the provider is located and the outer space where the rover is located. Figure 3 illustrates the sequence of interactions between the rover client and the rest of the entities.

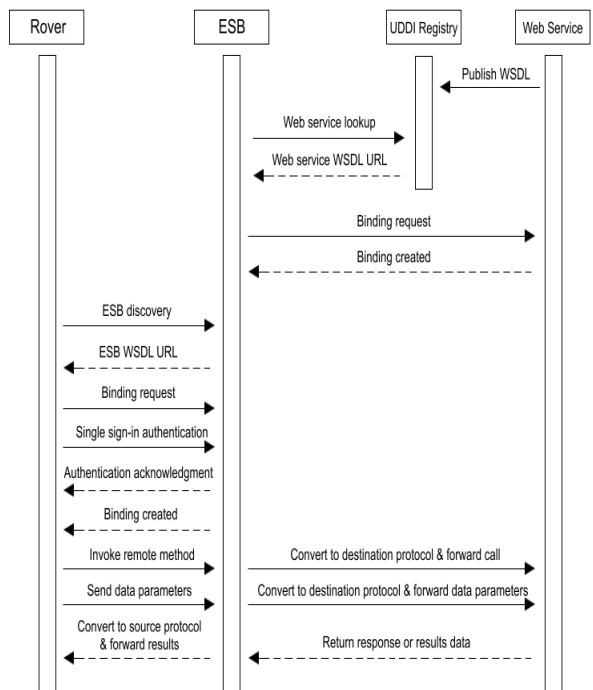

**Figure 3.** The Rover's Sequence of Interactions

### 5.2.2 The Middleware Tier – The ESB

The ESB or Enterprise Service Bus provides a data-path for data to travel between the rover unit and the different web services. It constitutes a data transmission medium, emulating a messaging middleware that links between the rover from one side and the different distributed services from the other side to allow them to send and receive data back and forth to each other. Additionally, it automates the in and out communications between all involved parties and coordinates the interaction between them, and allows the storage, routing, and transformation of messages during inter-system interactions.

The proposed ESB is cross-platform and cross-network which allows the rover to interoperate with various types of web services, possibly incompatible, built using different platforms, different standards, different technologies, and different programming languages to send requests, and receive responses from each other. Figure 4 depicts the architecture of the proposed ESB together with its inner-workings.

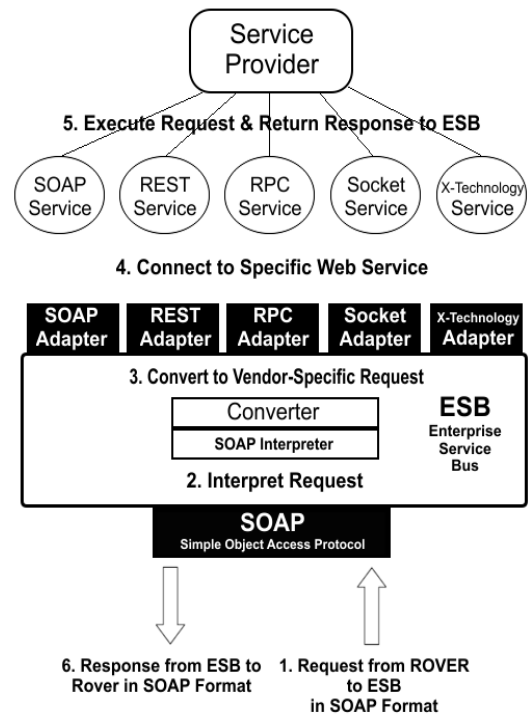

**Figure 4.** The Architecture of the proposed ESB

In effect, the ESB has two public interfaces: The first interface is from the rover's side which provides a unified single SOAP-based end-point for the rover to communicate with the ESB. The second interface is from the web services' side which provides a set of adapters as end-point connectors for the different web services to connect. There exists an adapter for every web service protocol, for instance, SOAP, REST,

RPC, Network Socket, and others. The role of these adapters is to bridge the rover's request to its destination web service irrespective of its protocol type and version. In order to achieve this, the ESB is able to identify the type of the request received from the rover and to route it accordingly to the corresponding adapter which, successively, passes it to the corresponding web service. All in all, the ESB delivers a transparent communication between the different components of the SOA allowing them to interoperate despite their underlying incompatible technologies and platforms. The ESB communication process can be described as below:

Step 1: The rover invokes a function called *feature_detection()* located in a REST-based web service. The request is always in SOAP protocol and encapsulates metadata describing the request message, including the source client, the destination service, the function to call, and a set of parameters.

Step 2: The ESB receives the request message in SOAP format; it first validates the correctness of its XML structure and then converted from SOAP format into the protocol of destination web service, in this case REST, using the protocol translator. The ESB uses an internal registry to lookup the technical details about the destination web service.

Step 3: The ESB routes the converted request to the adapter that is compatible with the addressed web service, in this case, the REST adapter.

Step 4: The adapter then locates the corresponding web service and gets bound temporary to it and starts executing the requested function, in this case *feature_detection()*.

Step 6: Once processing is done, a response is sent back from the destination web service to rover. It is first sent to the corresponding adapter, in this case, the REST adapter, then to the ESB, then translated to an SOAP format, and eventually routed to the rover.

*5.2.3 The Server Tier – The Web Services*

The server tier is where the web services are hosted. It mainly consists of several mainframe computer servers located either in earth control centers or in a space station in the low earth orbit. These servers define the execution of the web services, process rovers' requests, execute business logic, and perform intensive calculations on behalf of the rover. The web services can be of any type, protocol, or version and they interact with the ESB through its multi-platform end-point adapters. Each time a new web service is integrated into the system, it publishes its WSDL to the ESB which save it inside an internal registry along with other important details. The ESB then exposes the WSDL to the rover vehicle allowing it to call remotely all available functions.

Web services can provide any type of functionalities including computer vision functionalities to analyze captured images and recognize objects within these images; navigation functionalities to allow the rover to relocate and move over the surface of the planet; sensing functionalities to measure the atmospheric properties surrounding the rover; microscopy functionalities to analyze and inspect the nature of rocks and soils and their structure; and scanning functionalities to detect the presence of certain elements inside the planetary terrain [29].

## 6. Experiments & Implementations

As a proof of concept, a robotic rover simulation software was built capable of performing various actions. Besides, it is capable of sending requests to and reading results from the ESB using the SOAP protocol. The software is a regular standalone executable application built using C#.NET under the .NET Framework 4.0 and the MS Visual Studio 2010. Figure 5 depicts the GUI interface of the rover simulation software.

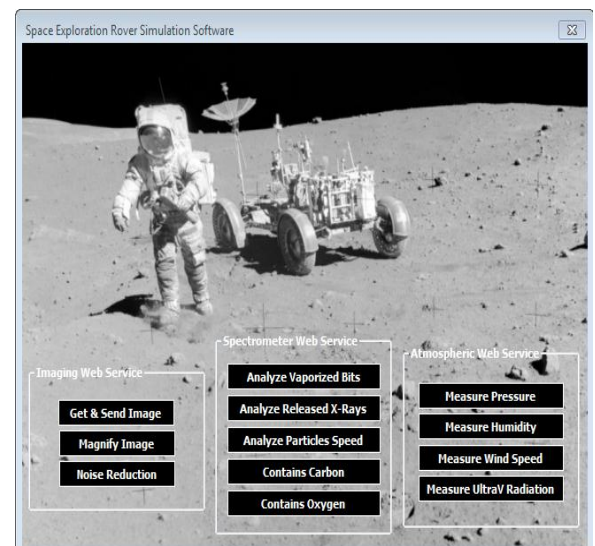

**Figure 5.** Rover's GUI interface

Additionally, three web services were developed. The first is a SOAP-based web service built using C#.NET with an *.asmx* file extension and is responsible for performing imaging functionalities including capturing color pictures of the planet landscape, and sending them to earth where scientists will study and analyze them. The second is a REST-based web service built using Java with a file extension *.jsp* and is responsible for performing spectrometer operations including mass spectrometer, gas chromatograph and laser spectrometer. Its purpose is to analyze rocks and soils in search for carbon, hydrogen, oxygen, and nitrogen-containing compounds. The third is a Socket-based web service built using C++ with an *.exe* file extension and is responsible for measuring atmospheric pressure, humidity, wind speed and direction, air temperature, ground temperature, and ultraviolet radiations. Figure 6 is a code snippet extracted from the source-code of the SOAP-based web service whose aim is to magnify any type of images that are captured by the rover's cameras.

```
[WebService(Namespace = "ImagingWebService",
 Description = "This Web Service Performs Various Imaging Manipulations")]
[WebServiceBinding(ConformsTo = WsiProfiles.BasicProfile1_1)]
public class ImagingWebService : System.Web.Services.WebService
{
    // <param name="Image">Image to Magnify</param>
    // <param name="newWidth">Image New Width</param>
    // <param name="newHeight">Image New Height</param>
    [WebMethod (Description="Zoom and Increase the Image Size")]
    public Bitmap MagnifyImage(Bitmap srcImage, int newWidth, int newHeight)
    {
        Bitmap newImage = new Bitmap(newWidth, newHeight);
        Graphics gr ;
        using (gr = Graphics.FromImage(newImage))
        {
            gr.SmoothingMode = SmoothingMode.AntiAlias;
            gr.InterpolationMode = InterpolationMode.HighQualityBicubic;
            gr.PixelOffsetMode = PixelOffsetMode.HighQuality;
            gr.DrawImage(srcImage, new Rectangle(0, 0, newWidth, newHeight));
        }
        return new Bitmap(newWidth, newHeight, gr);
    }
}
```

**Figure 6.** Magnify image method

Finally, an ESB was built to act as a middleware between the rover and the different web services. As the ESB is the service broker, it is responsible for exposing the different functionalities of the web services to the rover. Figure 7 delineates the list of functionalities exposed by the ESB and originally implemented in the web services.

**ESB**
This is the ESB - Enterprise Service Bus
The following operations are supported.

- **AnalyzeParticlesSpeed**
  Search for Water Minerals by Firing Beams of Neutrons and Measure Speed of Bounced Particles
- **AnalyzeReleasedXRays**
  Measure the Abundances of Various Chemical Elements by Shooting X-rays and Measuring their Release Amount
- **AnalyzeVaporizedBits**
  Measure the Composition of Rocks by Firing a laser at rocks and Analyzing the Structure of their Vaporized Bits
- **ContainsCarbon**
  Check if Underneath Soil Contains Carbon Element
- **ContainsOxygen**
  Check if Underneath Soil Contains Oxygen Element
- **MagnifyImage**
  Zoom and Increase the Image Size
- **MeasureHumidity**
  Measure the Atmospheric Humidity
- **MeasurePressure**
  Measure the Atmospheric Pressure
- **MeasureUltravioletRadiation**
  Measure the Atmospheric Ultraviolet Radiations
- **MeasureWindSpeed**
  Measure the Atmospheric Wind Speed
- **NoiseReduction**
  Filter and Correct Image Artifacts
- **SendImage**
  Send Captured Image to the ESB

**Figure 7.** Various methods exposed by the ESB

A use case scenario [30] was created for evaluation of the proposed model. Its purpose is to test the validity and the interoperability of the rover-web services communication through the ESB.

1. The rover simulation software needed to execute function *AnalyzeParticlesSpeed()* located in the REST-based web service, so it connected to the ESB in a process to discover all public available functionalities.
2. Once function *AnalyzeParticlesSpeed()* was found, the rover bound to the ESB and sent an authentication message to the ESB
3. The ESB acknowledged the rover allowing it to start remote function invocation.
4. The rover invoked function *AnalyzeParticlesSpeed()* sending mass=5 and weight=10 as parameters to the ESB using the SOAP protocol.
5. The ESB received the call and then looked-up for the destination web service that encapsulates function *AnalyzeParticlesSpeed()*.
6. Once the corresponding REST-based web service was found, the ESB converted the rover's call message from SOAP into REST and forwarded it to web service.
7. The REST-based web service received the call, it directly processed it, and execute function *AnalyzeParticlesSpeed()* on its hosting server.
8. Upon finishing processing, the web service returned the result velocity=11.332 to the ESB in REST format.
9. The ESB converted the REST message into a SOAP message readable by the rover, and forwarded it to it.
10. The rover received the results and displayed it on the screen.

Furthermore, other use cases were executed at runtime while the system was running and in all situations the rover succeeded to adapt itself according to the new changes in the environment. The different uses case scenarios are given below:

1. Integrating a new web service
2. Removing an existing web service
3. Updating web service functionalities
4. Failing an existing web service
5. Fixing a faulty web service
6. Deriving new web services from existing ones

## 7. Validation of the Proposed Architecture

The SOA approach proved to be very effective in all the different executed scenarios. The interoperability of the system allows the collaboration between various entities regardless of their underlying technologies and implementation details. The scalability of the system allows the support team to easily and quickly alter and add functionalities to the rover without having access to it. The maintainability of the system allows fixing or replacing out of order services while the system is running with no or minimal operation interruption. The reusability of the system allows building new web services or deriving new web services from existing ones with the least amount of development time and cost.

## 8. Conclusions & Future Work

This paper presented a novel architecture for building space exploration robotic rover systems using distributed software components called web services. The proposed architecture consists of three tiers: the client tier corresponding to the rover vehicle that requires executing some functionalities; the server tier corresponding to the web services that delivers the rover' functionalities; and the ESB acting as a middleware that coordinates and shields the complexity and heterogeneity of communication among the different entities of the system. Experiments conducted showed a robust, reliable, scalable, interoperable, reusable, and a maintainable architecture that can adapt itself to the unforeseen circumstances and cope with the various obstacles that might be encountered during real exploration missions.

As future work, the proposed SOA architecture is to be secured by adding to it an encryption layer which would protect and conceal the exchange of messages and data communication among the various entities of the system.

## Acknowledgments

This research was funded by the Lebanese Association for Computational Sciences (LACSC), Beirut, Lebanon under the "Service Oriented Architecture Robotics Research Project – SOARRP2012".